\documentclass[10pt,twocolumn,letterpaper]{article}

\usepackage{iccv}
\usepackage{times}
\usepackage{epsfig}
\usepackage{graphicx}
\usepackage{amsmath}
\usepackage{amssymb}
\usepackage{comment}
\usepackage{algorithm}
\usepackage{algorithmic}
\usepackage[accsupp]{axessibility}

\usepackage[pagebackref=true,breaklinks=true,letterpaper=true,colorlinks,bookmarks=false]{hyperref}

\iccvfinalcopy 

\def\iccvPaperID{5345} 

\begin{document}

\title{Benchmarking Algorithmic Bias in Face Recognition:\\ An Experimental Approach Using Synthetic Faces and Human Evaluation}

\author{Hao Liang\\
Rice University\\
{\tt\small hl106@rice.edu}
\and
Pietro Perona\\
California Institute of Technology and AWS\\
{\tt\small perona@caltech.edu, peronapp@amazon.com}
\and
Guha Balakrishnan\\
Rice University\\
{\tt\small guha@rice.edu}
}
\maketitle

\begin{abstract}
We propose an {\em experimental} method for measuring bias in face recognition systems. Existing methods to measure bias depend on benchmark datasets that are collected in the wild and annotated for protected (e.g., race, gender) and non-protected (e.g., pose, lighting) attributes. Such {\em observational} datasets only permit correlational conclusions, e.g., ``Algorithm A's accuracy is different on female and male faces in dataset X.''. By contrast, experimental methods manipulate attributes individually and thus permit {\em causal} conclusions, e.g., ``Algorithm A's accuracy is affected by gender and skin color.''

Our method is based on generating synthetic faces using a neural face generator, where each attribute of interest is modified independently while leaving all other attributes constant. Human observers crucially provide the ground truth on perceptual identity similarity between synthetic image pairs. We validate our method quantitatively by evaluating race and gender biases of three research-grade face recognition models. Our synthetic pipeline reveals that for these algorithms, accuracy is lower for Black and East Asian population subgroups. Our method can also quantify how perceptual changes in attributes affect face identity distances reported by these models. Our large synthetic dataset, consisting of 48,000 synthetic face image pairs (10,200 unique synthetic faces) and 555,000 human annotations (individual attributes and pairwise identity comparisons) is available to researchers in this important area. 

\end{abstract}

\section{Introduction}
Face recognition technology has applications in consumer media, information security, access control, law enforcement, and surveillance systems. The one-to-one task is to determine whether a pair of faces share the same identity (``face verification''). The one-to-many task determines whether a face image shares an identity with one or more from a database of face images from known individuals (``face identification''). Face recognition systems implemented with deep neural networks today achieve impressive accuracies~\cite{zhao2003face,deng2019arcface,liu2017sphereface,schroff2015facenet} and outperform even expert face analysts~\cite{phillips2018face}. Nevertheless, it is important to detect and measure possible algorithmic \emph{ biases}, i.e., systematic accuracy differences, especially across protected demographic attributes like age, race and gender~\cite{chiroro1995investigation,johnson2005we,hills2013eye}, in order to maintain fair treatment in sensitive applications.
For this reason, the National Institute of Standards and Technology (NIST) measures bias in commercial face recognition models~\cite{grother2019face}, in particular by comparing their False Match Rate (FMR) and False Non Match Rate (FNMR) values across different demographic subgroups at a particular decision threshold (sweeping this threshold yields FNMR vs. FMR ``curves'').

The first step in measuring bias of face recognition systems is, currently, to collect a large benchmarking dataset containing a set of diverse faces, where each is photographed multiple times under different conditions. An algorithm's error rate across subgroups specified by different protected attribute combinations (e.g., different race and gender groups) can then be measured. 

Unfortunately, sampling a good test dataset is almost impossible. First, each protected intersectional group (a specific combination of attribute values) must contain a sufficiently large number of individuals, which is not easy when sampling faces in the wild. Second, in order to estimate the \emph{causal} effect of these protected attributes on algorithm bias, it is crucial to ensure that the joint distribution of non-protected attributes like age, lighting, pose, etc. is roughly equal across protected attribute groups -- otherwise biases in the dataset will be misinterpreted as biases in the algorithm. It is practically impossible to do this with \emph{observational} data, i.e., data that we have no power to construct or intervene on ourselves, and all available datasets fall short of this criterion. Third, privacy concerns with collecting human data, and the cost and accuracy of identity annotation, make dataset construction for face recognition benchmarking very expensive. 



We address this challenge by developing an approach to construct \emph{synthetic} face recognition benchmarking datasets using a combination of modern face generation models and human annotators. We assume access to a pretrained face generator with a latent space, such as any of the popular public models trained in a generative adversarial network (GAN) framework \cite{goodfellow2020generative,lin2022raregan,karras2019style,Chan2021,Karras2019stylegan2}. We traverse the generator's latent space to first construct many ``identities'' (actually, ``pseudo-identities'' since there is no real person behind the face images) spanning different protected attributes (race and gender in our experiment). For each identity, we vary several non-protected attributes (pose, age, expression and lighting) to construct face sets that ideally depict the same person in different settings. We vary the non-protected attributes in the same way for all faces to ensure a balanced dataset. A challenge in using synthetic faces is we do not have a ground truth identity label per face. Such labels are needed to evaluate the accuracy of our algorithms. We replace the ground truth with consensus from human annotations, which we call the ``Human Consensus Identity Confidence'' (HCIC). We use the HCICs along with the synthetic images to benchmark face recognition systems for bias. 

Our approach is fast, practical, inexpensive, and virtually eliminates privacy concerns in data collection. The closest related work uses synthetic faces to benchmark face analysis systems~\cite{balakrishnan2021towards}, which classify attributes like gender and expression for a single face. We build on their ideas, but address the unique challenge of obtaining ground truth annotations for identity comparisons \emph{between face pairs}. Our work is the first to present an annotation pipeline for verifying a synthetic face recognition benchmarking dataset, and to demonstrate that a synthetic approach can be reliable for causal face recognition benchmarking.

Using our synthetic dataset consisting of 48,000 synthetic face image pairs (10,200 unique synthetic faces) and 555,000 human annotations (individual attributes and pairwise identity comparisons), we computed face identity distances reported by three public face recognition models. 
We used the HCICs to assign a ground truth to each pair, and computed False Non Match Rate (FNMR) vs. False Match Rate (FMR) curves for different attributes and demographic groups. Our results show that these algorithms have lowest error rates on White (a shorthand for European-looking, or Caucasian) groups.
For each model, we also report the expected change to predicted face identity distance with respect to changes to each nonprotected attribute. 

We make three contributions: (a) A method to {\em experimentally} estimate bias in face verification algorithms, and thus estimate the causal connection between attributes and bias, eliminating confusion between test set and algorithmic bias. (b) An empirical evaluation of our method, where we discover lower accuracy for Black and East Asian faces compared to Caucasian faces in three popular academic-grade face verification algorithms. (c) A large dataset of synthetic faces with human-collected ground truth.



\section{Related work}
{\bf Face recognition.}
Face recognition~\cite{zhao2003face,deng2019arcface,liu2017sphereface,schroff2015facenet} is one of the most successful applications of image analysis and understanding, and has been applied to many areas including information security, access control, law enforcement, and surveillance systems. State-of-the-art models all use deep neural networks~\cite{deng2019arcface,liu2017sphereface,schroff2015facenet,wen2021sphereface2}, and typically work by embedding each face into a latent space, and assigning an identity distance between a pair of faces by computing a distance (typically cosine distance) in that latent space. These models are trained over large-scale face datasets\cite{Parkhi15,liu2015faceattributes,an2021partial,deng2019lightweight}, most of which contain thousands of identities and multiple images per identity.

{\bf Fairness in computer vision.}
There is an increasing focus on measuring and mitigating biases of computer vision models and datasets~\cite{grother2019face,zietlow2022leveling,ramaswamy2021fair,quadrianto2019discovering,du2020fairness,kearns2018preventing,torralba2011unbiased}. Biases may be measured with a number of metrics~\cite{vasudevan2020lift,hardt2016equality,corbett2018measure} to quantify performance differences of algorithms across population subgroups. Face recognition and analysis systems are often under the most scrutiny due to their sensitive nature~\cite{buolamwini2018gender, balakrishnan2021towards, karkkainen2021fairface,zietlow2022leveling}. A typical approach to measure bias of a face analysis algorithm is to collect a real dataset of faces with appropriate labels of protected attributes like race and gender, and compute model error across intersection subgroups of these attributes~\cite{buolamwini2018gender,albiero2021gendered,albiero2020does,de2021fairness,krishnapriya2020issues,lu2019experimental}. Recent work argues that measuring bias on datasets that are collected in the wild risks confusing dataset bias with algorithmic bias -- the fix is using an experimental approach using synthetic face generators~\cite{balakrishnan2021towards}.  As an added bonus using synthetic images lowers the cost and privacy concerns of data collection. We build on that study in this work, by extending their ideas to benchmark face recognition models.
 
{\bf Bias mitigation.} Methods for mitigating bias mainly focus on two elements: the training dataset and the model design. A dataset has sampling biases if its joint distribution of attributes is far from random. For example, the CelebA face dataset is known to have significant sampling biases, such as a higher proportion of female with young ages compared to men~\cite{balakrishnan2021towards}. A model trained on a biased dataset may inherit biases~\cite{karkkainen2021fairface,zemel2013learning,ponce2006dataset}, particularly for attribute subgroups that are underrepresented in the dataset. Conversely, bias can be mitigated if the algorithm is trained on a balanced dataset. Researchers are also building novel model designs to combat biases\cite{zhang2018mitigating,zou2018ai,karras2020analyzing,wang2019racial,wang2019balanced}, typically by learning representations invariant to protected attributes or sampling in a balanced way during training. 

{\bf Face generation.} 
Several algorithms can produce near photorealistic synthetic faces: Variational Autoencoders(VAE)\cite{kingma2013auto,van2017neural,liang2020controlled}, GANs~\cite{goodfellow2020generative,lin2022raregan,karras2019style,Chan2021}, and diffusion models~\cite{ho2020denoising,nichol2021improved}. 
We use StyleGAN~\cite{karras2019style, Karras2019stylegan2,Chan2021}. A useful property of a GAN is that it generates images from a low-dimensional latent space where it is often possible to find a one-dimensional subspace that alters a given scalar attribute like gender or age. The most common strategy is to train a linear classifier or regressor that predicts values of an attribute from the latent vector, and then traverse along the norm of the resulting model's hyperplane to generate new images varying that attribute\cite{shen2020interpreting,harkonen2020ganspace,richardson2021encoding}. Higher-order methods that examine local latent space geometry and nonlinear traversals have also been proposed\cite{balakrishnan2022rayleigh}. 
\section{Method}
\begin{figure}[!tp]
    \centering
    {\label{fig:1}\includegraphics[width=0.48\textwidth]{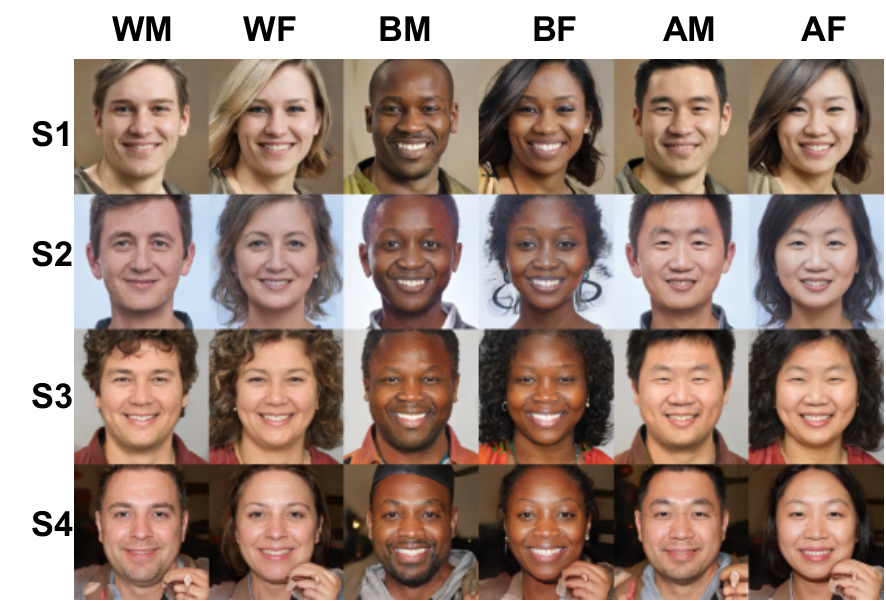}}
    \caption{\textbf{Examples of prototypes image spanning protected attributes of race and gender.} In the first step of our framework (see Sec.~\ref{method:sec1}), we generate random ``seed'' images by randomly sampling in the latent space of our face generator. Starting from each seed, we perform latent space traversals (see Sec.~\ref{method:sec1}) along latent dimensions correlated with race and gender to produce faces depicting different ``demographic groups" (M=Male, F=Female, W=White, B=Black, A=East Asian). Each row in this figure shows the six prototypes generated from the same seed image. Each column shows members of the same {\em group} that were generated from different {\em seeds}. See Fig.~\ref{fig:changing attributes} for the next step.}
    \label{fig:prototypes} 
\end{figure}

\begin{figure}[!tp]
    {\label{fig:1}\includegraphics[width=0.45\textwidth]{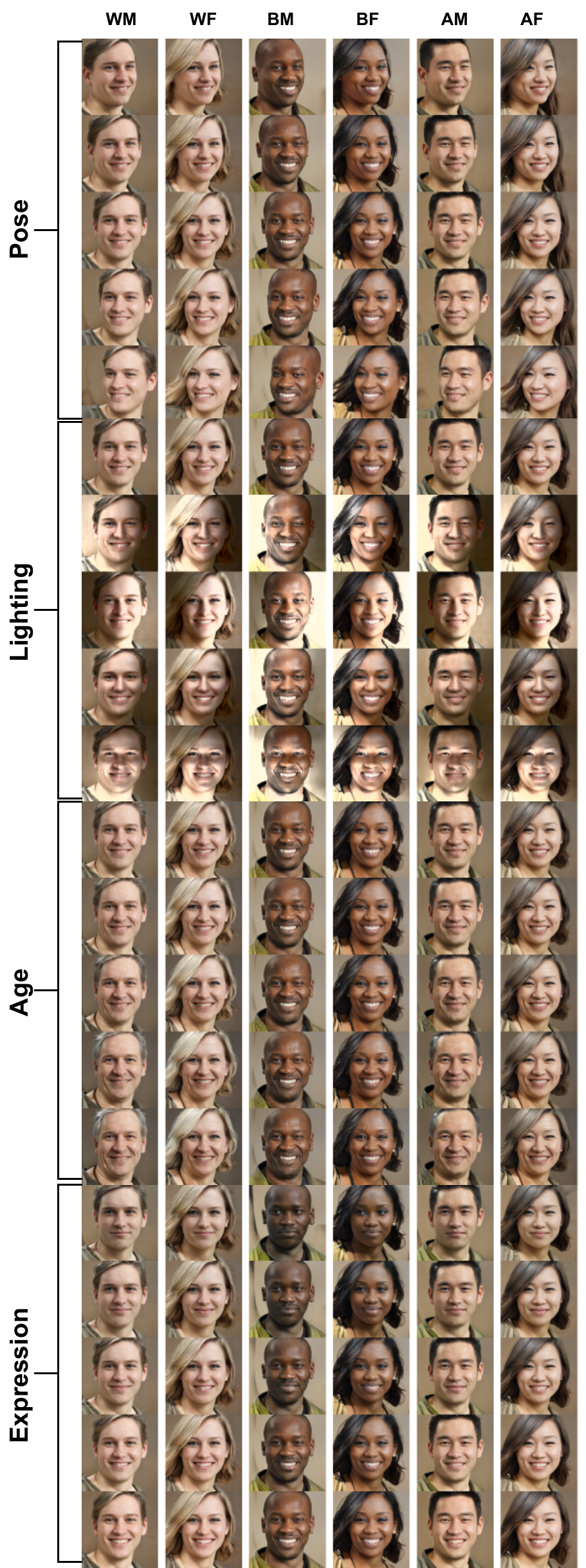}}
    \caption{\textbf{Examples of modifying non-protected attributes.} In the second step of our framework (see Sec.~\ref{method:sec2}), we modify each prototype face (see Sec. ~\ref{method:sec1} and Fig.~\ref{fig:prototypes}, 1$^{st}$ row) along attributes. All 120 images here come from one seed (S1 in Fig.~\ref{fig:prototypes}). Each column (20 images) is generated from the same prototype.}
    \label{fig:changing attributes}
\end{figure}

Our method consists of seven steps. First, generate {\em seed } face images (see Fig.~\ref{fig:prototypes}) by sampling the latent space of the GAN. Second, generate {\em prototype} face images spanning all protected attribute combinations (race and gender in our experiments) by controlling features in the latent space of a face generator (see Fig.~\ref{fig:prototypes}) (we use the abbreviations: \{``WM'', ``WF'', ``BM'', ``BF'', ``AM'', ``AF''\} for \{``White Male'', ``White Female'', ``Black Male'', ``Black Female'', ``East Asian Male'', ``East Asian Female''\}). Third, generate systematic modifications of each prototype image along non-protected semantic attributes (pose, age, expression, and lighting) (see Fig.~\ref{fig:changing attributes}). These form sets of faces that correspond to the same intended identity. Fourth, form pairs of images, both within and across identity sets. Fifth, give face pairs to human annotators to obtain perceived identity match scores which we call ``Human Consensus Identity Confidences'' (HCICs). Sixth, ask human annotators to determine the attributes of single images and the similarity of image pairs. Seventh, feed the synthetic pairs to face recognition models, and use the HCICs to evaluate them for biases. We describe these steps in the next sections.

\subsection{Generating images of different demographic groups along protected attributes}
\label{method:sec1}
We assume a pretrained face generator that maps vectors $\mathbf{z} \sim p(z) \in R^d$ to images, where $R^d$ is a low-dimensional {\em latent space} and $p$ is a given probability density.  In this work, we use a generator trained in a GAN framework. In particular, we use EG3D~ \cite{Chan2021}, a state-of-the-art generator with explicit control over geometry and pose.

We aim to generate a large set of ``prototype'' faces that depict all combinations of protected attribute intersectional subgroups. We consider the protected attributes of race (limited for simplicity to the East Asian, Black, Caucasian subgroups) and gender (similarly limited to the Male, and Female subgroups), 
resulting in 2 (genders) $\times$ 3 (races) = 6 prototypes generated per seed. By using one seed to generate each set of 6 prototypes, our intention is to keep attributes other than race and gender approximately constant, so that the six seeds are {\em matched} vis-a-vis potential confound variables, which is useful for later statistical analysis.

Our method for modifying the seed images to produce prototypes is based on linear latent space traversals \cite{balakrishnan2021towards,harkonen2020ganspace,richardson2021encoding,shen2020interpreting}. We sample a large number of images from $p(z)$, and label race and gender of each using a public multi-task classifier~\cite{karkkainen2021fairface}. This produces a training set $D = \{I^j, Z^j, L^j \}_{j=1}^{N}$, where $I^j, Z^j,L^j$ denotes the face image, latent code, and labels of sample j respectively, and $N$ is the total number of samples (100,000 in our experiments). Second, we train one support vector machine (SVM) to classify gender (assuming binary labels), and three one-vs-all SVMs for each of the three race subgroups. For a seed image, we generate a set of prototypes by starting at the seed's position in latent space and moving along the normal directions of each attribute's SVM hyperplane.

Not all generated images are realistic, and some of the seeds are similar in appearance to each other. To eliminate unrealistic images and quasi-duplicates, we filter the many random seeds into a smaller set that has diverse and realistic prototypes. We accomplish this by selecting 300 (out of 1000 initial) seeds using a human annotator judging the realism and agreement of the associated prototypes with respect to intended race and genders. Then, we further filter these 300 seeds into 100 using a max-min clustering algorithm (see Algorithm \ref{alg:algorithm} in Supplementary) evaluated over the geometry of the faces as captured by a public 3D face mesh predictor \cite{kartynnik2019real}. This yields 600 total prototype images and a total of 10,200 images (17 per prototype, Fig.~\ref{fig:changing attributes} and Sec.~\ref{method:sec2}).

\subsection{Modifying non-protected attributes}
\label{method:sec2}
After creating prototype images of different demographic groups, we modify them according to various semantic attributes to build our dataset. We modify the following attributes: pose, lightning, age, and expression. We adopt different techniques when manipulating the attributes, as detailed in the following subsections. Starting from each prototype, we generate one image sequence per attribute depicting different gradations of change to that attribute. Examples of these sequences are shown in Fig.~\ref{fig:changing attributes}. In our experiments, we set all sequence lengths to 5.

{\bf Pose:} Pose is explicitly modeled by the EG3D generator. 
For each prototype, we generate a sequence of images at angles of $\{-30^{\circ}, -15^{\circ}, 0^{\circ}, 15^{\circ}, 30^{\circ}\}$.

{\bf Lighting:} We manipulate lighting using a public neural network person relighting model~\cite{zhou2019deep}. We used four lighting conditions, produced by a light source located $\{up, down, left, right\}$ with respect to the face. We set the power of the light source to 0.7, yielding a strong, but not overwhelming lighting effect.

{\bf Age and expression:} We control these with the latent space traversal method mentioned in Sec.~\ref{method:sec1}. We use pretrained models to assign age and expression labels to all face images in $D$, and train linear regressors to predict these labels from the latent vectors. To generate the sequences, we start at a prototype's position in latent space and move along the normal vector of the regression hyperplane with a small step size until a certain score (0.8 for age, 0.9 for expression after normalizing the outputs of models to $[0,1]$) is predicted by the model at a distance $d$ from the prototype. Then, we take $n$ steps from the prototype of size $d/n$ along both directions. We set $n=2$ in our experiments, resulting in $2n + 1 \text{(prototype)} = 5$ images per image sequence.


\subsection{Face segmentation and background removal}
We further white-out all pixels not corresponding to the head (background, shoulders, neck) using a public face segmentation model~\cite{yu2018bisenet}. This is for two reasons. First, some images have artifacts outside of the face (see Fig\ref{fig:prototypes} seed 4, for artifacts of fingers). Although most face recognition models first run face detection to crop regions outside of the face, we still remove the background for safety. Second, the background may inadvertently affect human annotator judgments. An example image before and after face segmentation and background removal is in Fig.~\ref{fig:bgrm} in Supplementary.

\begin{figure*}[ht!]
    \centering
    {\label{fig:1}\includegraphics[width=1\textwidth]{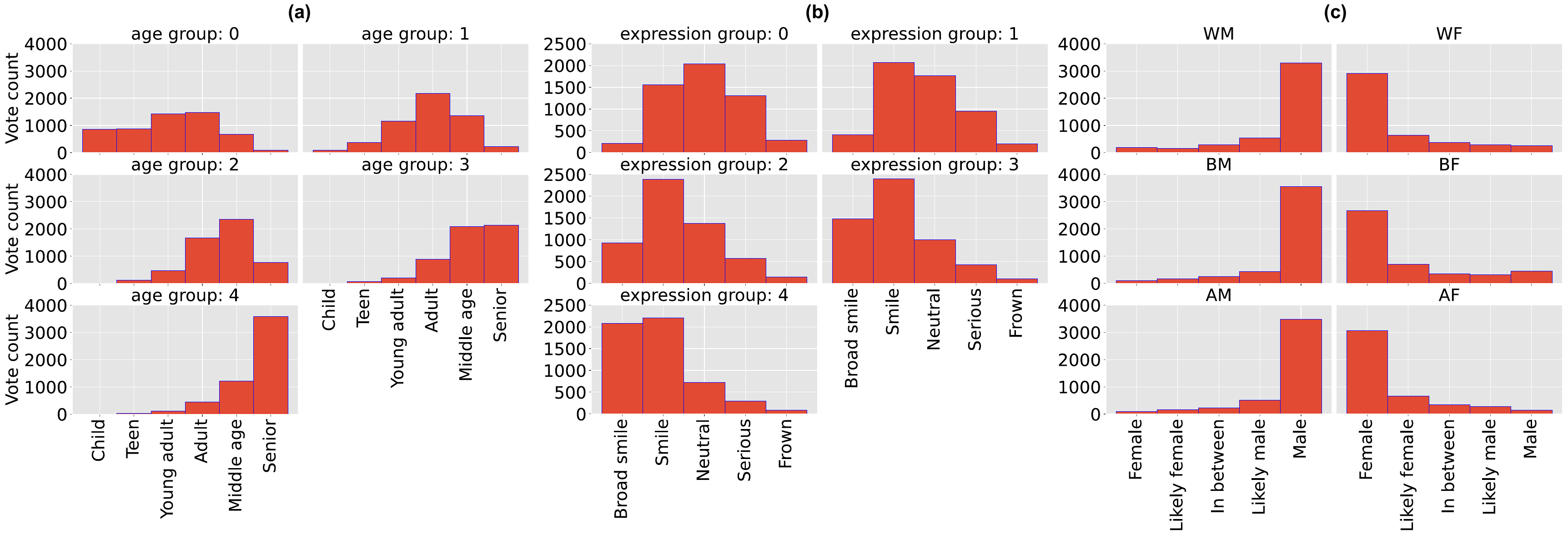}}
    \caption{\textbf{Validation of attribute manipulation through single image annotation}.  \textbf{(a)}: Annotator scores for age, ranging from ``child" to ``senior", on single synthetic images belonging to age groups 0 (youngest-looking) to 4 (oldest-looking) (Fig.~\ref{fig:changing attributes}).  \textbf{(b)}: Annotator scores for expression, ranging from ``frown" to ``broad smile", each histogram is associated to a different expression group, from 0 (most frowning) to 4 (most smiling).  \textbf{(c)}: Annotator scores ranging from ``female" to ``male". Histograms are shown for different demographic groups. The annotators' perceptions track along with our manipulations and validate our method. See Fig.~\ref{fig:single} in Supplementary for results on skin tone and uncanny.
    }
    \label{fig:single}
\end{figure*}

\subsection{Generating face pairs}
\label{Sec:gen}
A face recognition test dataset consists of both `positive' and `negative' face pairs, where positive pairs depict the same identity (see Fig.~\ref{fig:changing attributes}) and negatives do not. 
We create positive pairs by taking each prototype face and pairing it with all other faces we generated starting from the prototype, varying one non-protected attribute at a time (see Fig.~\ref{fig:changing attributes}). 
We construct negative pairs by pairing each prototype image with faces from sequences produced by $n$ \emph{other} prototypes of the {\em same race and gender}, also varying one non-protected attribute at a time. In this way, a negative face pair will depict a change to a single attribute, along with a highly likely change in identity. Because the number of total possible negative pairs produced in this manner is large, we set $n=3$ in our experiments. Therefore, our database consists of $12,000$ positive 
pairs and $36,000$ negative pairs.

\subsection{Human annotation experiments}
\label{Sec:human}
The attribute modifications we perform in Secs.~\ref{method:sec1} and~\ref{method:sec2} may not always work as intended. In particular, attributes can change at different rates in different sections of the generator's latent space. More importantly, we have no ground truth measurements for face identity similarity between image pairs. To address both issues, we collect human annotations using {\em Amazon SageMaker Ground Truth}. For each image we collected 9 human annotations per attribute and average their responses to obtain one score.

{\bf Single image annotation:} We give annotators \textbf{one} image from the synthetic dataset at a time, and ask them to annotate the skin type, perceived gender, expression, age, and uncanniness (fakeness) using five-point scales. See Fig.~\ref{fig:interface} in Supplementary for screenshots of our annotation interfaces.

{\bf Image pair annotation:} 
To ensure that a pair of faces \textit{\textbf{does}} belong to same/different person(s) from the perspective of a majority of humans, we collect pairwise ground truth annotations. For each pair of images 9 annotators are asked to choose from the following options regarding identity: \textit{\{‘likely same’, ‘possibly same’, ‘not sure’, ‘possibly different’ and ‘likely different’\}}. A screenshot of the annotation interface is in Fig.~\ref{fig:interface} in Supplementary.



\section{Experiments}
We evaluate our method on three public, research-grade face recognition models: a ResNet-34 trained on Glint360k using Arcface~\cite{deng2019arcface}, a ResNet-34 trained on MS1MV3 using Arcface~\cite{deng2019arcface} and a SFNet-20 trained on VGGFace2 using Sphereface\cite{liu2017sphereface}. All of the models were trained on large in-the-wild datasets and reached high accuracy on their respective test datasets. During inference, for a pair of face images$(I_A, I_B)$, each model returns feature vectors $(f_A, f_B)$, and similarity between faces is quantified using cosine similarity between the vectors: $(f_A \cdot f_B) / (||f_A|| ||f_B||)$.

Following the image generation pipeline introduced in Sec. \ref{method:sec1}, \ref{method:sec2}, we build our synthetic dataset consisting of $10,200$ face images depicting $600$ intended identities, and further construct $12,000$ positive face pairs, and $36,000$ negative face pairs following the generation pipeline proposed in Sec. \ref{Sec:gen}. We obtained 9 human annotations per image and per attribute to quantify the nonprotected attributes of age and expression (we did not annotate pose and lighting as our method of controlling them is much more precise), as well as 9 annotations per image pair to annotate identity distances. In total, we collected 123,000 annotations from 2,214 annotators (median=65, max=223 annotations each) for single image attribute annotations and 432,000 annotations from 1,905 annotators (median=264, max=402 annotations each) for image pair annotations.

\begin{figure*}[ht!]
    \centering
    {\label{fig:1}\includegraphics[width=\linewidth]{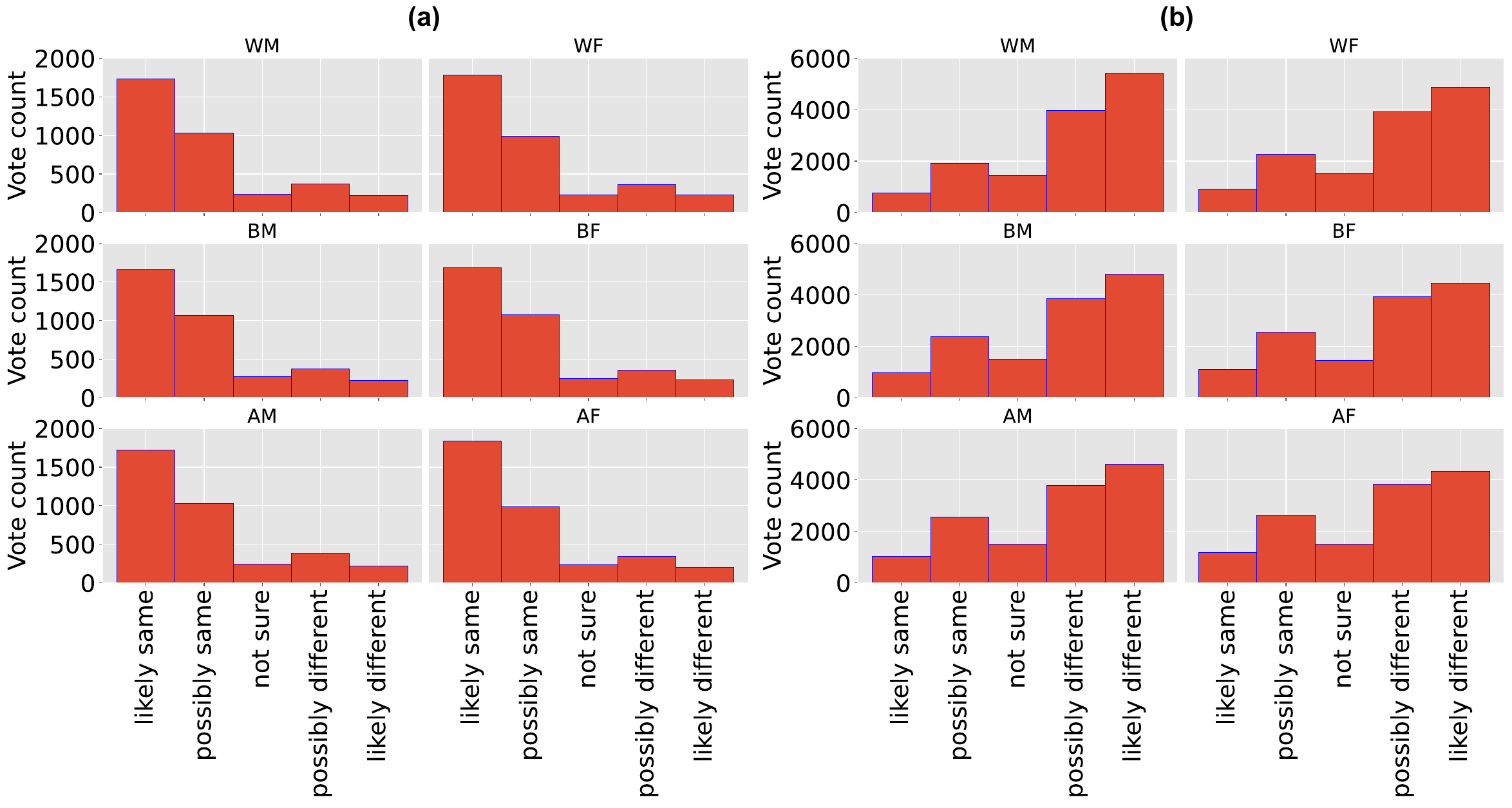}}
 
    \caption{\textbf{Human annotation results on different pose image pairs.} \textbf{(a):} face images of different poses from the same seed (putative identity) and same demographic group. \textbf{(b):} face images of different poses from different seed (different putative identy) and same demographic group. Crucially, we do not find statistical differences in the similarity or difference of face pairs across different demographic groups. This is consistent across attributes, see Fig.~\ref{fig:model_eval1} in Supplementary for results on other attributes.}
    \vspace{-1em}
    \label{fig:pose human}
\end{figure*}
\begin{figure*}[!t]
    \centering
    {\label{fig:1}\includegraphics[width=1\textwidth]{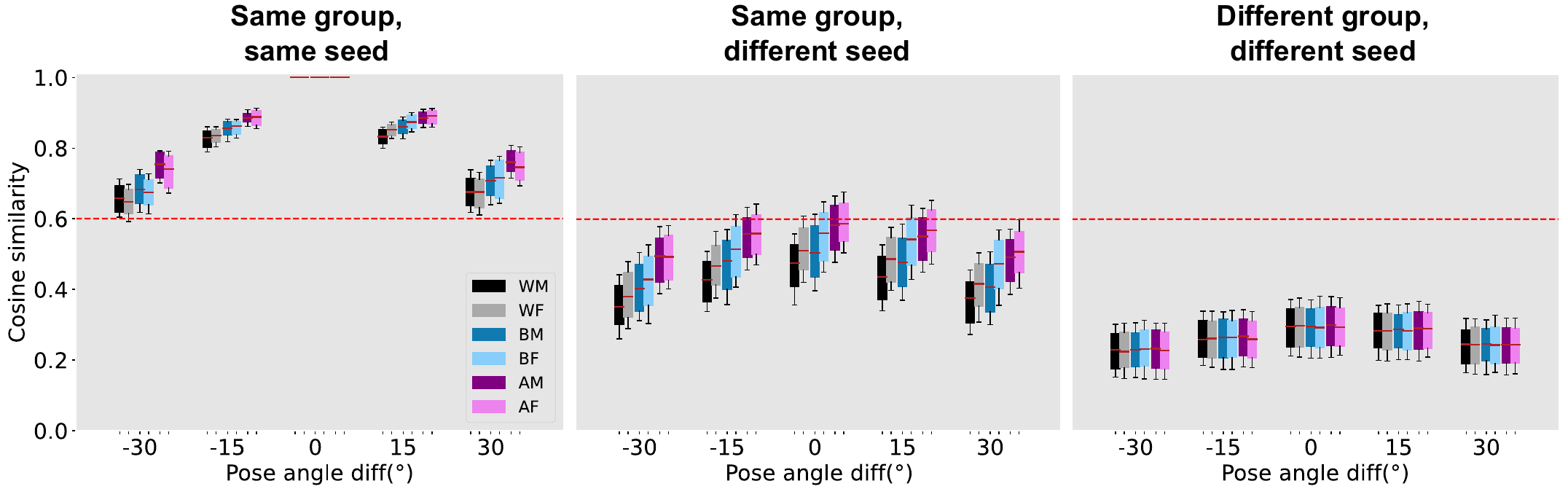}}
    \caption{\textbf{Identity similarity scores reported by a recognition model with respect to pose and demographic changes}. As described in Sec.~\ref{section:algeval}, we feed face image pairs of different groups (see Fig.~\ref{fig:prototypes}) to a public recognition model. Box plots show the median cosine distance (red line in the boxes) and the $15\%-85\%$ confidence interval. The red dashed line at 0.6 indicates a reasonable threshold $t$ to decide whether a pair of images are from the same person or not (the threshold trades off False Non Match Rate (FNMR) for False Match Rate (FMR) errors). The left plot corresponds to face pairs from the same seed and demographic group, i.e., nominally the same `individual.' The second corresponds to face pairs from the same demographic group and different seed (i.e., different individuals with the same gender and similar racially-correlated attributes), and the last plot corresponds to face images from different demographic groups and seeds. It is clear that race and gender have a strong effect on the perception of identity.
    Results for other attributes are in Fig.~\ref{fig:model_eval1} in Supplementary.}
    \label{fig:model_eval}
\end{figure*}

\begin{figure*}[ht!]
{\label{fig:1}\includegraphics[width=1\textwidth,height=6in]{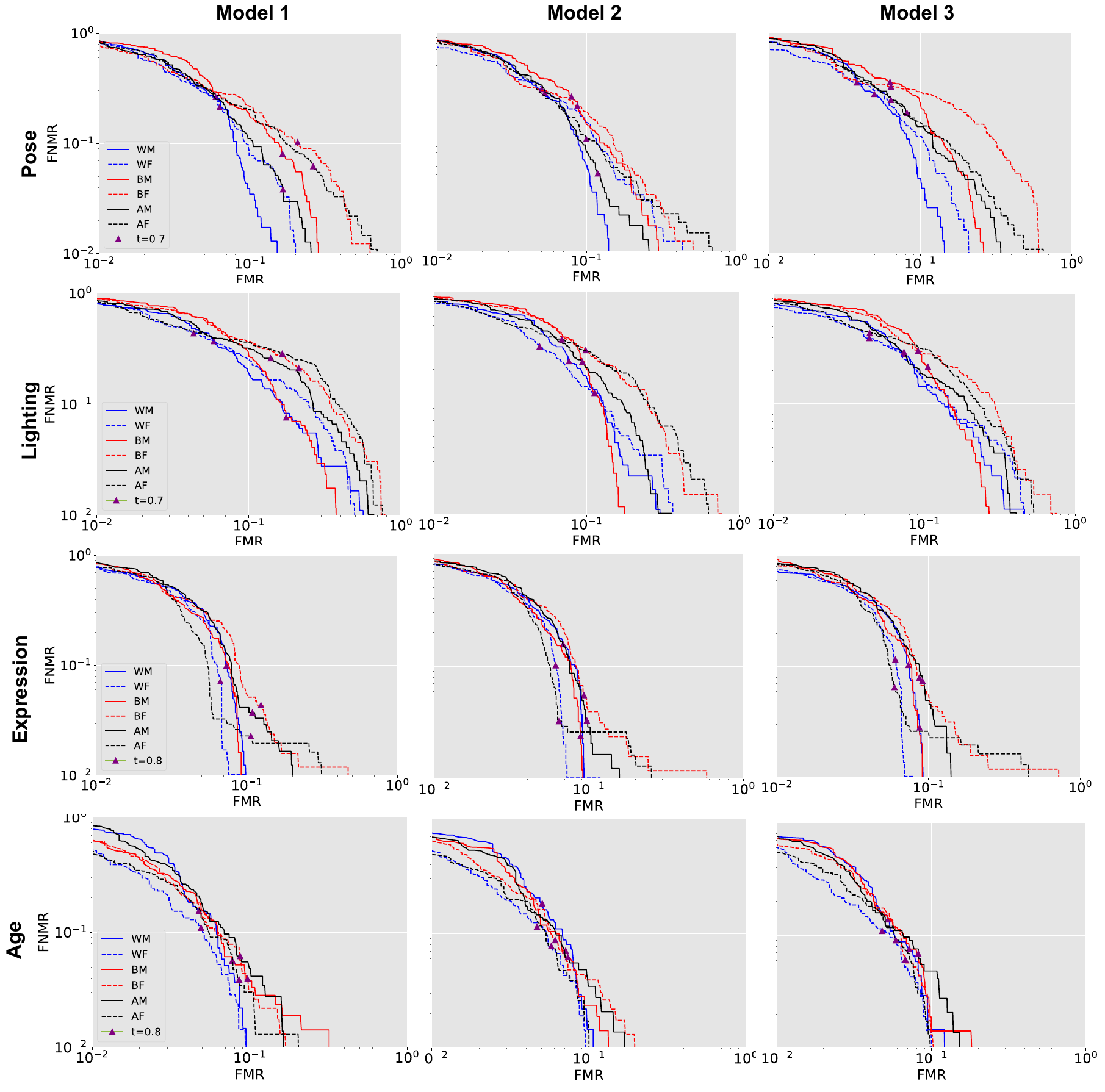}}
    \caption{\textbf{FNMR vs. FMR plots} from all image pairs using HCIC with $t_{hcic} = 0.3$ as the ground truth labels (see Sec.~\ref{section:algeval}). Each row shows FNMR vs. FMR vis-a-vis changes in one attribute: pose, lighting expression and age.  Each column refers to one of three models: model 1 is a SFNet trained on VGGFace2 using Sphereface~\cite{liu2017sphereface}, model 2 is a ResNet34 trained on MS1MV3 using Arcface~\cite{deng2019arcface}, and model 3 is a ResNet34 trained on Glint360k using Arcface~\cite{deng2019arcface}. 
    See Sec. \ref{section:algeval} for analysis on the results. Similar to ~\cite{grother2019face}, we use purple triangles to show how a {\it fixed} threshold $t$ (see caption of Fig. \ref{fig:model_eval} for details of $t$) can affect the performances of different demographic groups in different ways. Different choices on $t_{hcic}$ will not affect the overall trend, see Fig.~\ref{fig:fnmr2} \ref{fig:fnmr4} in Supplementary for results on different $t_{hcic}$ values.}
    \label{fig:fnmr}
\end{figure*}

\subsection{Annotation and image synthesis analysis}
Single image annotations for the attributes of age, skin color and gender are shown in Fig.~\ref{fig:single}.  Please see the caption of Fig.~\ref{fig:single} for details. The results show that annotator perceptions agree well with our intended manipulations across all race and gender subgroups.
We also asked annotators to label image realism (``uncanniness'' in our surveys), allowing us to remove any unrealistic or artifact-heavy images. More than $70\%$ of images receive a score of ``Likely real'' or ``real for sure'' across all demographic groups. See Fig.~\ref{fig:single} in Supplementary for details. 

Next, we evaluated the humans' identity distance annotations with respect to each non-protected attribute. Fig.~\ref{fig:pose human} shows distributions of the annotations for pose (plots of remaining attributes are in Fig.~\ref{fig:model_eval1} in Supplementary), broken down by race/gender subgroups and positive (a) and negative (b) face pairs. The annotators generally agreed with our intended synthesis, but there are also a significant number of images where they did not. These may be caused either by errors in synthesis or natural human perceptual variability. We show examples of image pairs for which annotators disagreed with our intention in Fig.~\ref{fig:bins} in Supplementary. 

Human annotator identity evaluations per image were also fairly consistent. See Fig.~\ref{fig:std} in Supplementary for a plot of the median standard deviations and $15\% - 85\%$ confidence interval for all attributes. All medians are roughly 0.3 (scale of $[0,1]$), indicating good consistency among annotators.

\subsubsection{Bias analysis of algorithms}
\label{section:algeval}

In this section, we report our main experimental results: bias analysis of the face recognition algorithms. We only use synthetic face pairs where each face receives an uncanniness score below 0.8, which results in $11,682$ positive face pairs, $35,406$ negative face pairs. Please refer to Fig.~\ref{fig:fail} in Supplementary for examples of face images which receive a score higher than $0.8$. 

We first evaluate the effect of changes to each attribute on each model's identity similarity prediction. Fig.~\ref{fig:model_eval} presents our results for the pose attribute, on the ResNet34-MS1MV3 model (see Fig.~\ref{fig:model_eval1}, \ref{fig:model_eval2}, \ref{fig:model_eval3} in Supplementary for plots of other attributes and models). As expected, the model reports highest similarity between faces from the same prototype (same demographic group, same seed), and least similarity for faces from different groups and seeds. In addition, we see a decline in reported similarity as the pose angle moves away from $0^{\circ}$, indicating that pose has a causal effect on the similarity score. This result also confirms another hypothesis: by comparing the second and third figure in Fig.~\ref{fig:model_eval}, it is straightforward to draw the conclusion that the model uses demographic group as essential information when making predictions. Other attributes also have similar causal effects on similarity score, and these effects are also consistent across test on different face recognition models, please see Fig.~\ref{fig:model_eval2}, \ref{fig:model_eval3} in Supplementary for details.

We also report False Non Match Rate (FNMR) vs. False Match Rate (FMR) for all models with respect to changes in their decision threshold in Fig.~\ref{fig:fnmr}, broken down by non-protected attributes. FNMR and FNR are calculated as: $\text{FNMR} = \frac{|\text{false reject pairs}|}{|\text{positive pairs}|}$ and $\text{FMR} = \frac{|\text{false match pairs}|}{|\text{negative pairs}|}$. As we vary the decision threshold of the face recognition model, we get a different number of false reject/false match pairs to draw the curves in the figure. 

To decide whether a given pair of images belongs to positive or negative class, we estimate the ground truth label from the human annotation scores. First, we compute the HCIC for each image pair, we remove the top two and bottom two scores from the the 9 human annotation scores. and compute the average score of the rest five. Next, we choose a threshold $t_{hcic}$ and compare the HCIC with it: we assign a positive label to pairs for which HCIC $\leq t_{hcic}$, and negative otherwise. 

Fig.~\ref{fig:fnmr} shows that the models have some clear biases. All models have lowest errors for the White Male and White Female groups. Results on pose changes indicate that all 3 models have significant biases towards race, with Model 3 performing significantly worse on Black Females. For lighting changes, model performances are reasonably consistent across demographic groups, though Model 2 clearly performs worse on Black Female and Asian Female. For expression, all three models again show obvious biases as they all perform best on White Males. For age changes, Model 1 is again significantly better on the White group, and all three models perform the worst on Black Males. Model biases with respect to gender are less consistent, though clearly present for some scenarios.


\section{Discussion and Conclusions}
We presented the first experimental approach to measure face recognition algorithmic bias based on generating synthetic images with attributes modified independently, and an identity ground truth computed from human annotator consensus. Our synthetic test dataset is obtained by generating 100 six-face balanced "prototypes" expressing two genders and three races, and then systematically varying non-protected attributes (pose, lighting, expression and age). Our final synthetic dataset is extensive, consisting of 12,000/36,000 positive/negative face pairs, and over 500,000 human annotations for both single image attributes and pairwise identity comparisons.

We validated our method by assessing three popular public face recognition models for bias. A first finding is that race and gender affect recognition algorithms deeply: it is much more difficult to recognize identity in tests within the same race and gender groups, then in test where faces belong to different groups. A second finding is that all models are biased: accuracy is better for White Males and Females and (in Model 3) lower for Black Females. One last observation is that that face recognition algorithms are more affected by variations in pose and expression, and less so by age and lighting.

Human annotations are consistent and reproducible (See Fig.~\ref{fig:std} in Supplementary). They confirm our attribute manipulations (single attribute annotations (Fig.~\ref{fig:single})), and show that perceived identity is preserved for same-prototype manipulations (identity comparisons between face pairs (Fig.~\ref{fig:pose human})). Furthermore, the accuracy of these annotators are consistent across race and gender subgroups (See Fig.~\ref{fig:model_eval} in Supplementary). This suggests that human perception can provide a robust and reproducible ground truth with which we may benchmark algorithms.



Our work use human annotations to estimate face identity, and the annotation process may be further improved. First, we could not ensure that our annotators come from different regions of the world. It is well known that humans are better able to identify others within their own demographic group than outside of it. While we do not see demographic biases in our annotation histograms, it would be good to make sure that annotators come from diverse demographics. Furthermore, it will be interesting to measure annotator response to same/different faces in images of real people and compare these responses to our synthetic image measurements. It is also possible that annotators may latch onto image factors that we did not foresee to make their judgments, such as when two faces generated in the same latent traversal have near-identical hairstyles.  Hairstyle does affect face recognition in most models, but SOTA models may be able to ignore hair. We have explored this by using BiSeNet segmentation \cite{yu2018bisenet} to remove hair and obtained similar trends to our current results.

Our generation process also has limitations. While the method based on latent space disentanglement has provided a simple and effective way to modify attributes, it can also generate artifacts or change other attributes in addition to the one(s) we intended (See Fig.~\ref{fig:fail} in Supplementary). This is exacerbated when trying to modify multiple attributes at a time~\cite{balakrishnan2021towards}. A valuable future direction would be to thoroughly compare different state-of-the-art generative models (GANs and diffusion models in particular) in terms of their ability to precisely control one or more face attributes. 




{\small
\bibliographystyle{ieee_fullname}
\bibliography{egbib}
}

\onecolumn
\appendix
\begin{center}
\textbf{\large Supplemental Materials}
\end{center}

\section{Max-min clustering algorithm for seeds filtering}
We use an algorithm based on max-min clustering as described in Alg. \ref{alg:algorithm} to filter the seeds. Refer to Sec. 3.1 for more details.
\begin{algorithm}
\caption{Max-min clustering seeds filtering}
\label{alg:algorithm}
\begin{algorithmic} 
\REQUIRE Base seeds $S_B$, filtered seeds $S_F$, face mesh model $M$, seeds requied $n$, DemographicGroup(DG) = \{``WM'', ``WF'', ``BM'', ``BF'', ``AM'', ``AF''\}
\STATE $S_F \leftarrow \{S_B[0]\}$
\WHILE{$i \leq n$}
\FOR{$s \in S_B \setminus S_F$} 
\STATE $D(s) = \min_{\forall f \in S_F, \forall g \in DG} = ||M(I_s^g) - M(I_f^g)||_2$
\ENDFOR
\STATE $S_F$.add$(\operatorname*{arg\,max}_s D(s))$
\STATE $i = i+1$
\ENDWHILE
\end{algorithmic}
\end{algorithm}

\section{Image annotation interface for identity comparison}
\begin{figure*}[h]
    \centering
    {\label{fig:1}\includegraphics[width=0.8\textwidth]{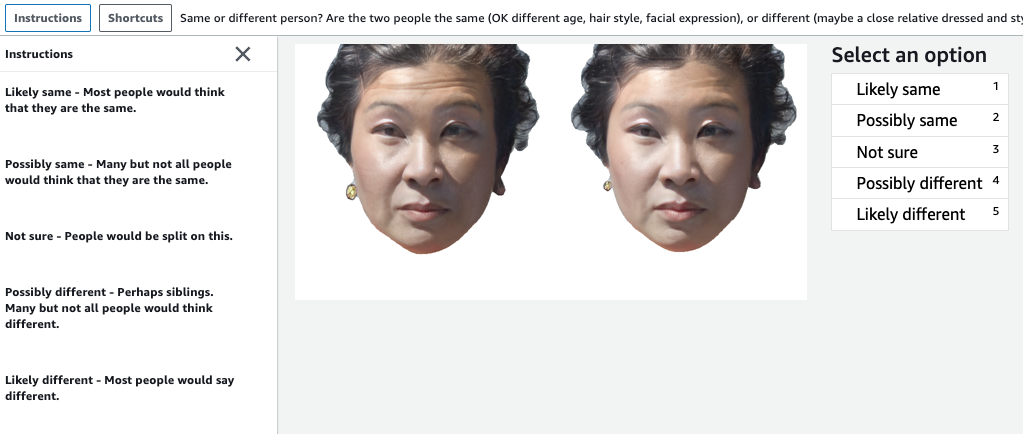}}
    \caption{\textbf{Human annotation interface.} We give annotators a pair of face images and ask them to choose from one of the following options: \textit{\{‘likely same’, ‘possibly same’, ‘not sure’, ‘possibly different’ and ‘likely different’\}}.}
    \label{fig:interface}
\end{figure*}

We collect human annotations using \textit{Amazon SageMaker Ground Truth} and we show an example of interface in Fig.~\ref{fig:interface}. Refer to Sec. 3.5 for details.

\clearpage
\section{Example image before \& after face segmentation and background removal}
\begin{figure}[!hbtp]
    \centering
    {\label{fig:1}\includegraphics[width=0.5\textwidth]{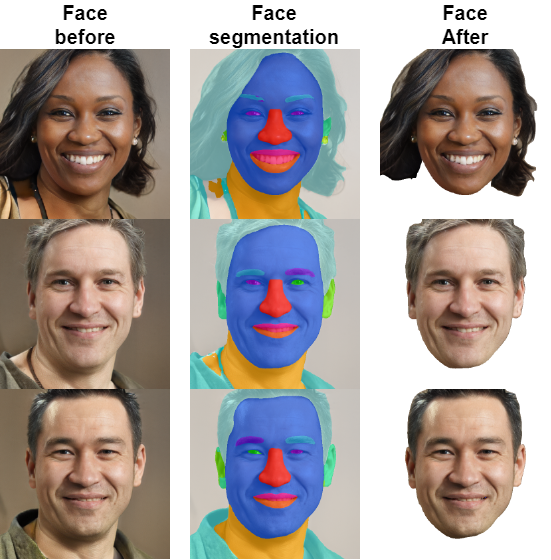}}
    \caption{\textbf{Face segmentation \& background removal}. To reduce noise in both model and human's prediction, we perform face segmentation and background removal to all the images.}
    \label{fig:bgrm}
\end{figure}

We show examples of images before and after segmentation and background removal in Fig.~\ref{fig:bgrm}. Refer to Sec. 3.3 for details.
\clearpage

\section{Single image annotation results for skin tone and uncanny.}
We asked annotators to label image realism (“uncanniness ” in our surveys) as well as race (``skin color'' in our surveys), the results are in Fig.~\ref{fig:single}. This allows us to throw away any images that are unrealistic or have significant artifacts. Refer to Sec. 4.1 for details.

\begin{figure}[H]
    \centering
    {\label{fig:1}\includegraphics[width=1\textwidth]{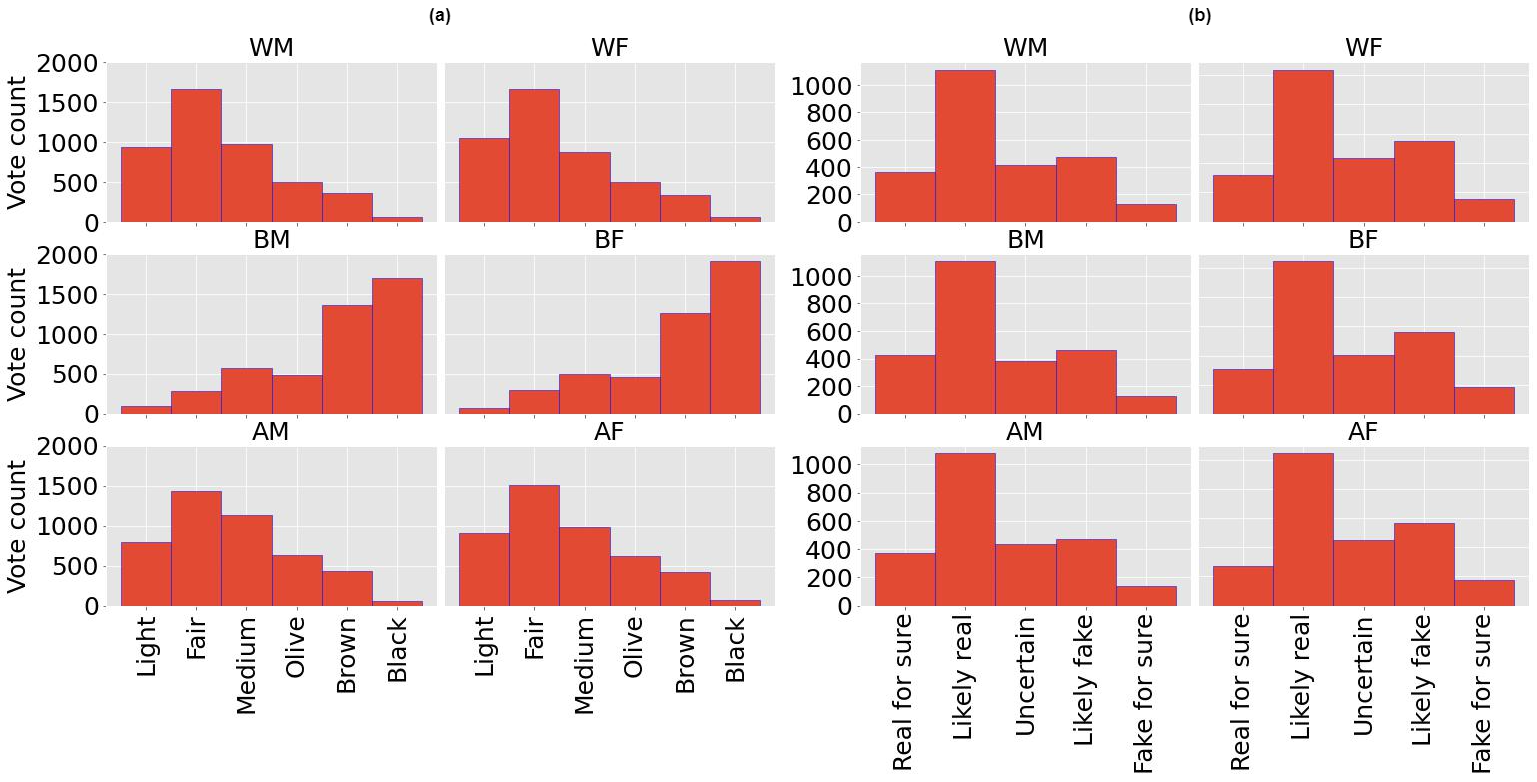}}
    \caption{\textbf{Single image annotation}.  \textbf{(a)}: Results for skin color, where annotators give scores per image ranging from "light" to "black". \textbf{(b)}: Results for uncanniness, where annotators give score per image ranging from "real for sure" to "fake for sure". We split results by demographic groups.
    }
    \label{fig:single}
\end{figure}

\clearpage
\section{Identity similarity scores for other attributes and models}
\begin{figure*}[!hbtp]
    \centering
    {\label{fig:1}\includegraphics[width=1\textwidth]{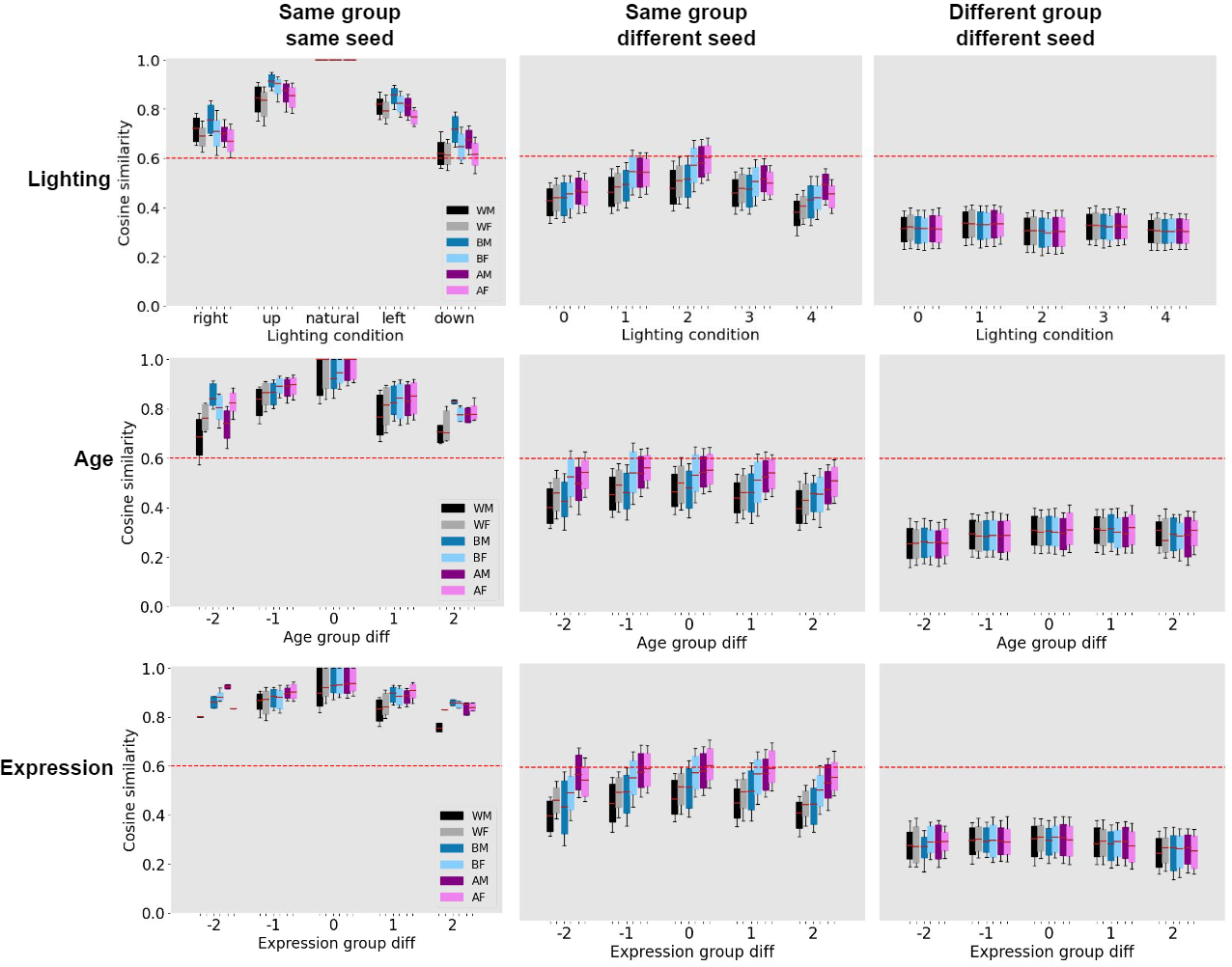}}
    \caption{\textbf{Identity similarity scores reported by ResNet34 trained using ArcFace on the MS1MV3 dataset with respect to non-protected attributes and demographic changes}.}
    \label{fig:model_eval1}
\end{figure*}

\begin{figure*}[!hbtp]
    \centering
    {\label{fig:1}\includegraphics[width=1\textwidth]{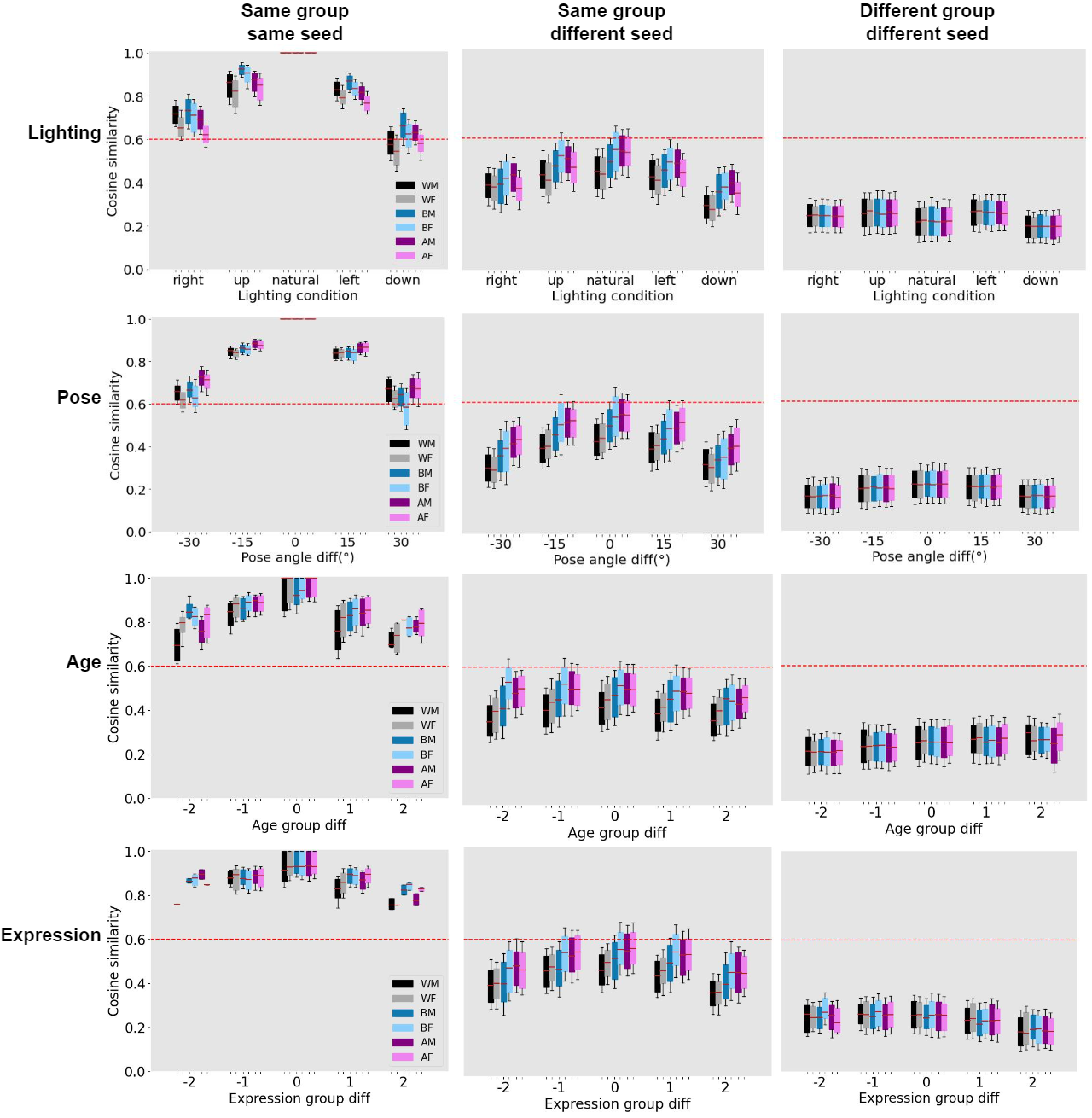}}
    \caption{\textbf{Identity similarity scores reported by ResNet34 trained using ArcFace on the Glint360k dataset with respect to non-protected attributes and demographic changes}.}
    \label{fig:model_eval2}
\end{figure*}

\begin{figure*}[!hbtp]
    \centering
    {\label{fig:1}\includegraphics[width=1\textwidth]{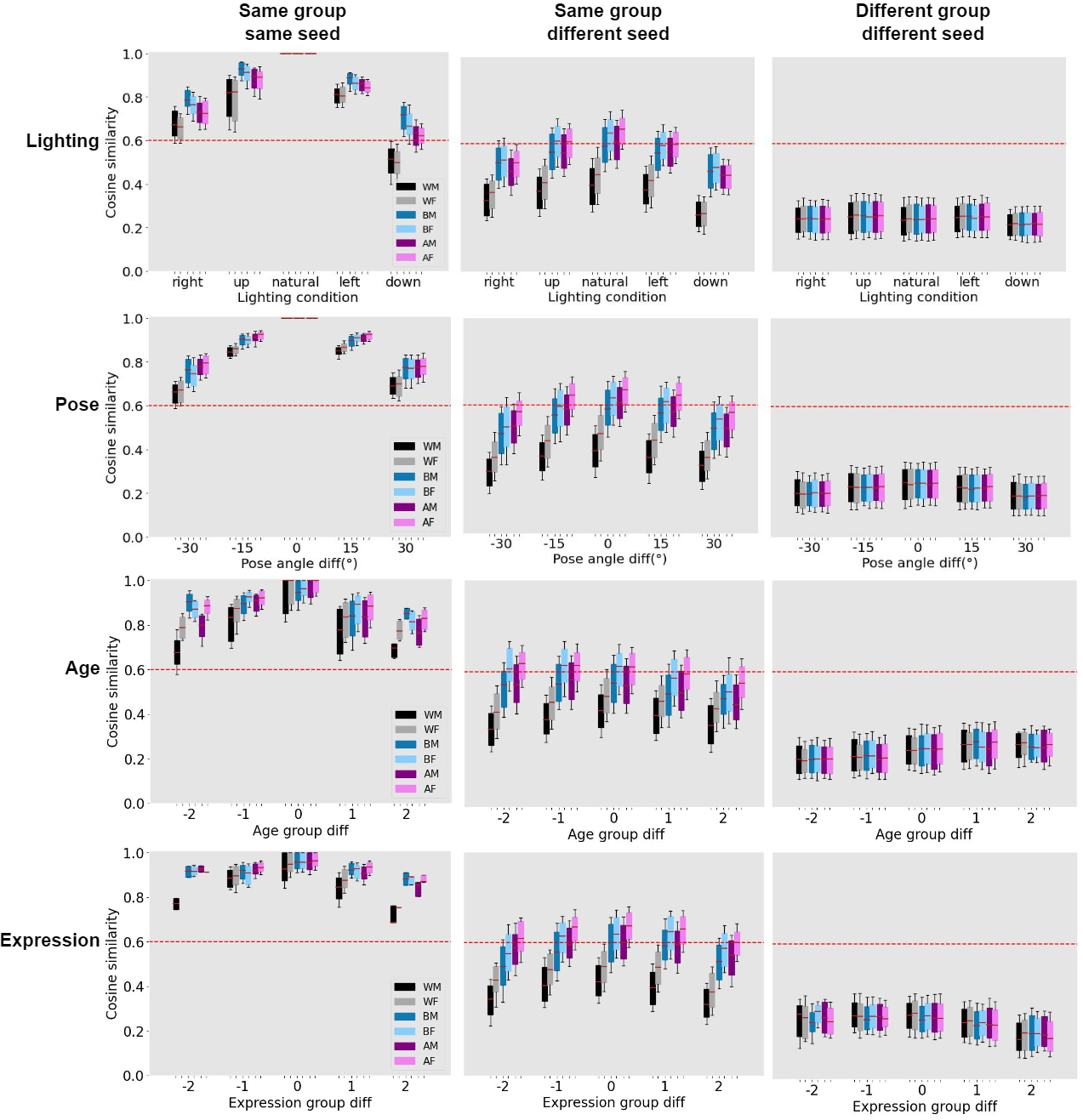}}
    \caption{\textbf{Identity similarity scores reported by SFNet20 trained using SphereFace on the VGGFace2 dataset with respect to  non-protected attributes and demographic changes}.}
    \label{fig:model_eval3}
\end{figure*}

We feed face image pairs to three different pretrained popular public face recognition models: a ResNet34 trained on MS1MV3 using ArcFace, a ResNet34 trained on Glint360k using ArcFace, and a SFNet20 trained on VGGFace2 using SphereFace. Note that since we don't have ground truth labels for ``age'' and ``expression'', we instead use the results from single image annotation(see Sec. 4.1) to assign them age/expression group: group $\{0,1,2,3,4\}$ represent images whose scores are in $\{[0,0.8), [0.8,1.6), [1.6,2.4), [2.4,3.2), [3.2,4]\}$ respectively. The results are shown in Fig \ref{fig:model_eval1}, \ref{fig:model_eval2}, \ref{fig:model_eval3}. Refer to Sec. 4.1.1 for details. 

\clearpage
\section{Per-Image Standard Deviations of Human Annotations}
\begin{figure}[!hbtp]
    \centering
    
    {\label{fig:1}\includegraphics[width=0.5\textwidth]{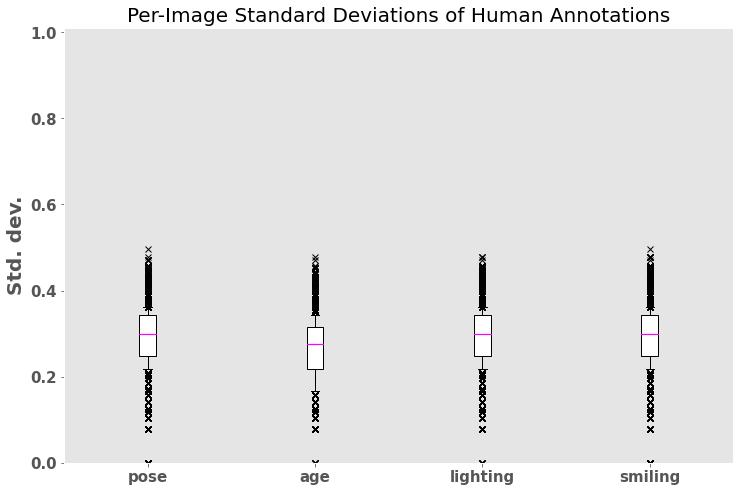}}
    \caption{\textbf{Per-Image Standard Deviations of Human Annotations}. Distributions of per-image standard deviations of human annotations for each of the attributes we considered (one unit = dynamic range of the attribute). Nine annotators were asked to provide a rating for an face image pair of each image. The median standard deviations are plotted in red lines, all of the medians are around 0.3, indicating good consistency among annotators.}
    \label{fig:std}
\end{figure}

\clearpage
\section{Human annotation results for different attributes}
\begin{figure*}[!hbtp]
    \centering
    {\label{fig:1}\includegraphics[width=0.85\textwidth]{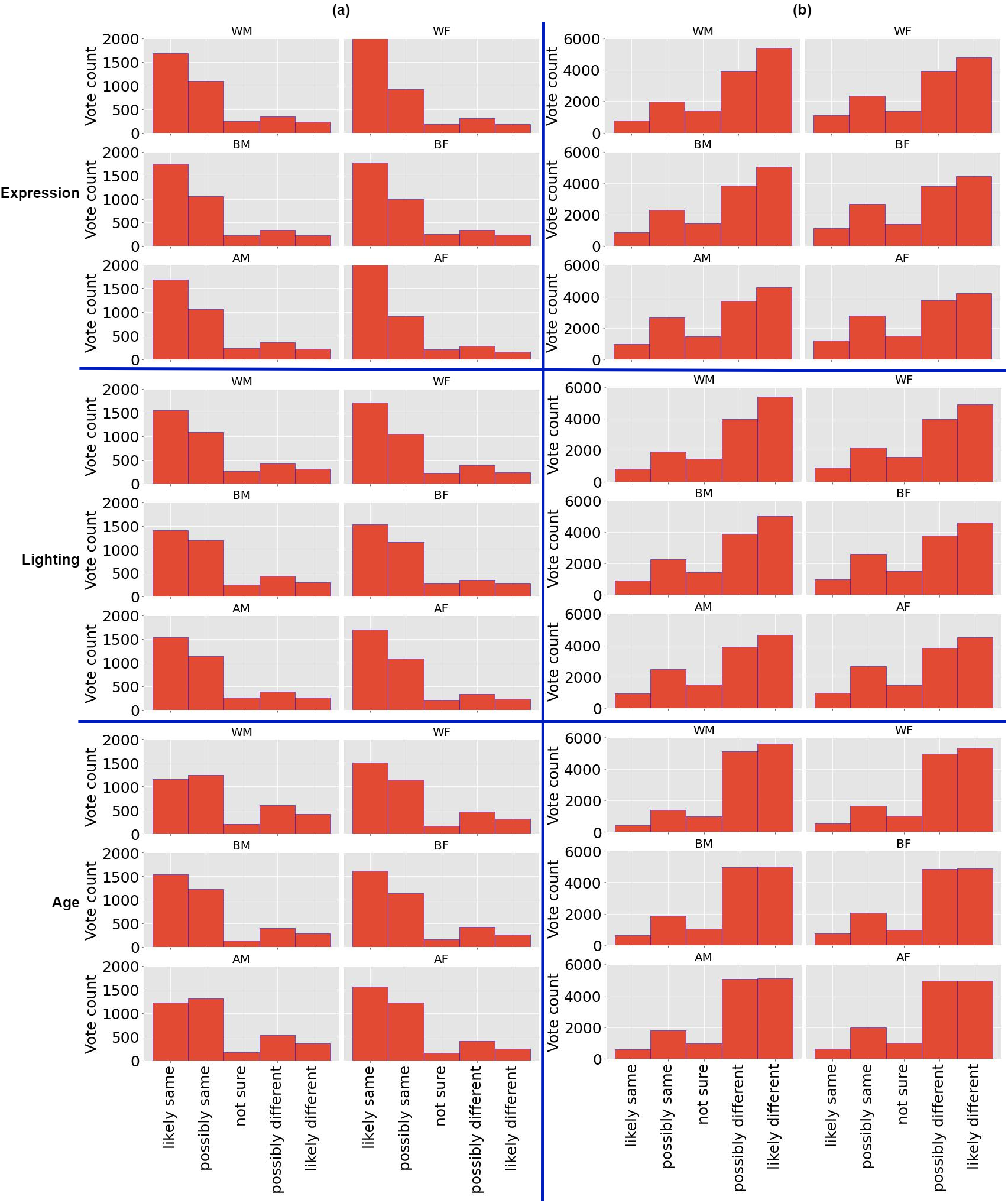}}
    \caption{\textbf{Human annotation results on different attribute image pairs.} \textbf{(a)} Image pairs from same seed and same demographic group. \textbf{(b)} Image pairs from different seeds but same demographic group. We see a consistent trend across all attributes. Refer to Sec. 4.2 for details.}
    \label{fig:model_eval4}
\end{figure*}
\clearpage

\section{FNMR vs. FMR results for different $t_{hcic}$ values}
\begin{figure*}[!hbtp]
  {\label{fig:1}\includegraphics[width=1\textwidth]{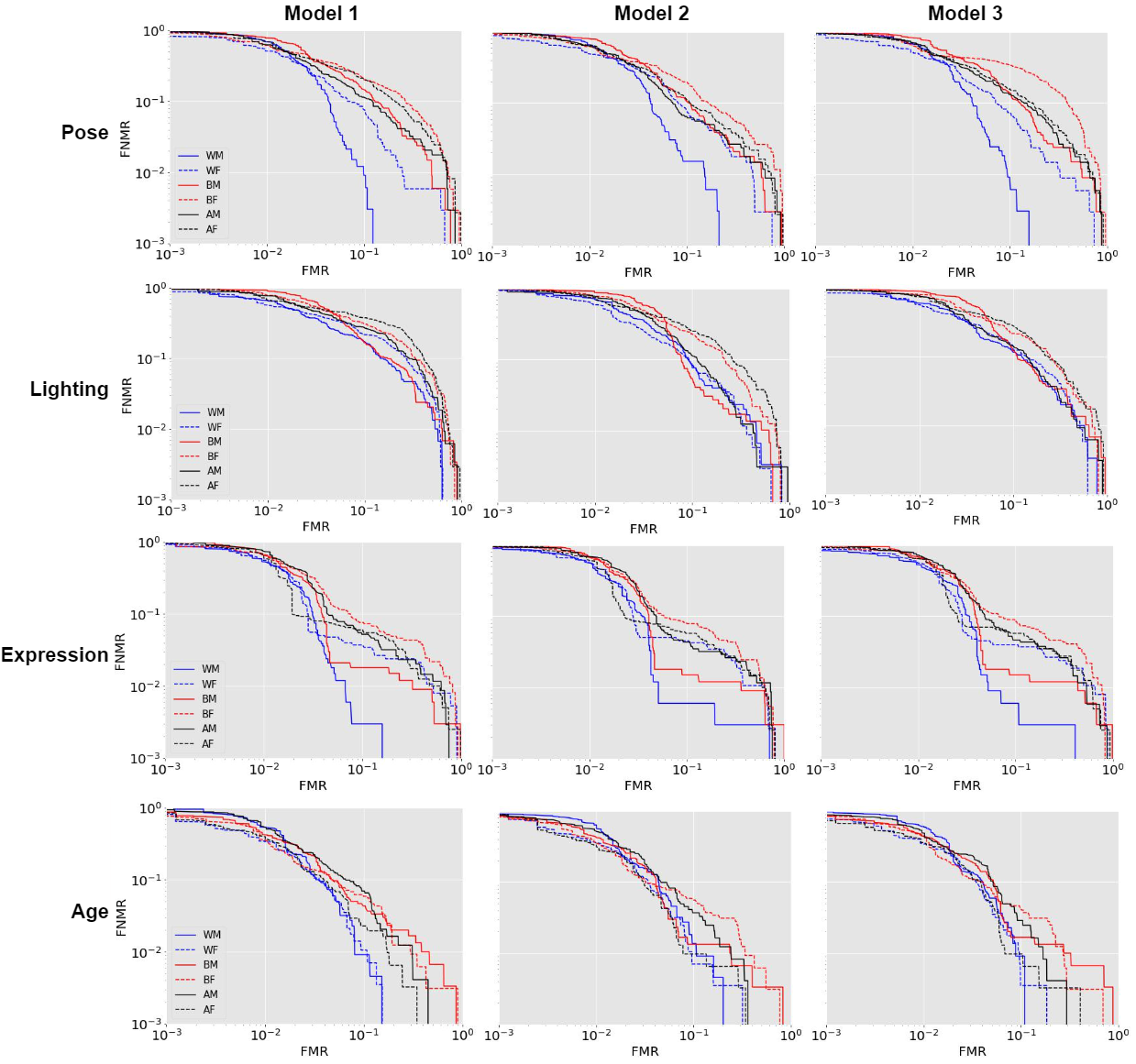}}
    \caption{\textbf{FNMR vs. FMR plots} from all image pairs using HCIC with $t_{hcic}=0.4$ as the ground truth labels.}
    \label{fig:fnmr2}
\end{figure*}

\begin{figure*}[!hbtp]
  {\label{fig:1}\includegraphics[width=1\textwidth]{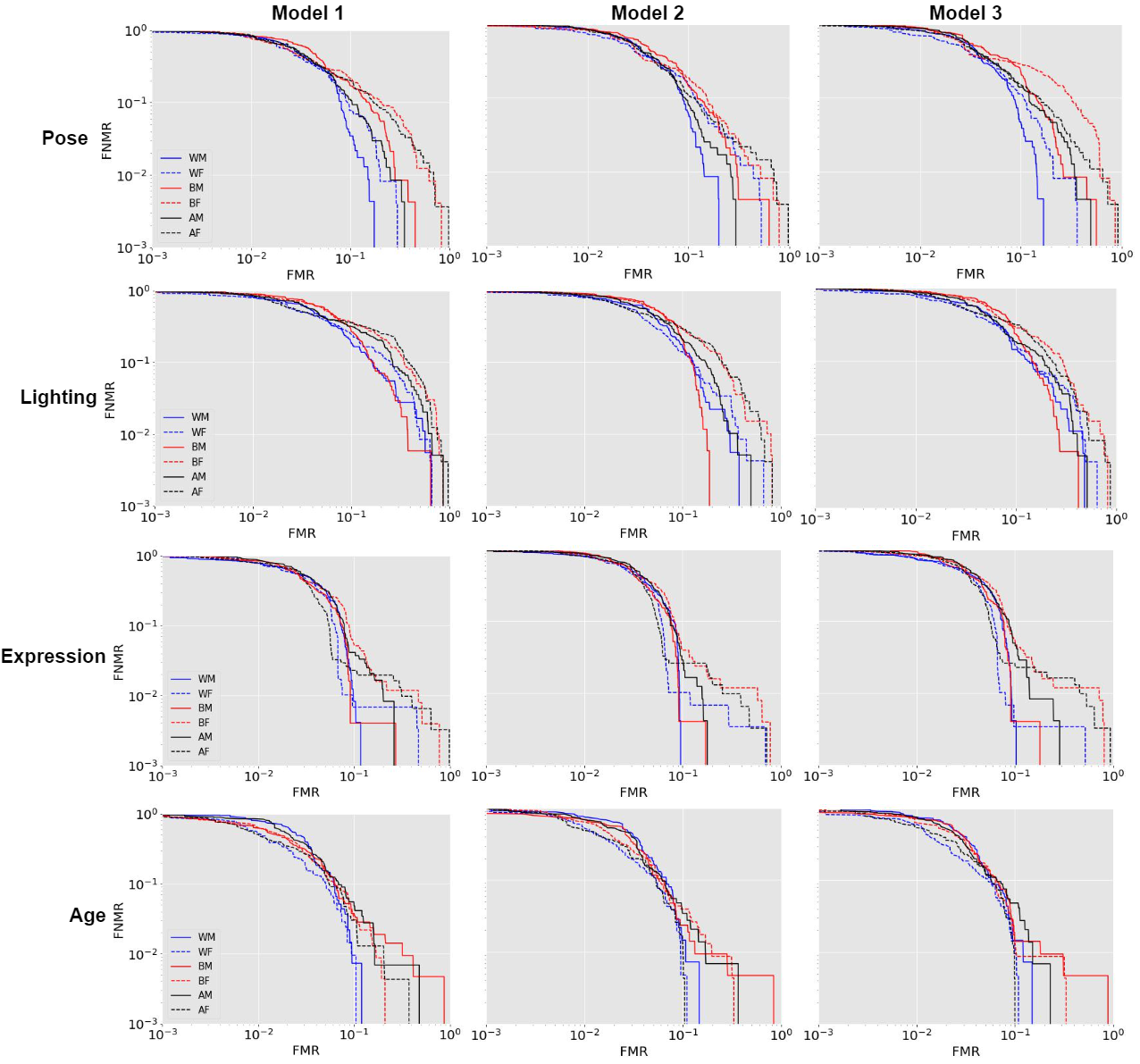}}
    \caption{\textbf{FNMR vs. FMR plots} from all image pairs using HCIC with $t_{hcic}=0.2$ as the ground truth labels.}
    \label{fig:fnmr4}
\end{figure*}

We show results of the FNMR vs. FMR plot with thresholds $t_{hcic} \in \{0.2, 0.4\}$ in Fig.~\ref{fig:fnmr2}, \ref{fig:fnmr4}. They basically show the same trend as $t_{hcic}=0.3$. Refer to Sec. 4.1.1 for details.

\clearpage
\section{More examples \& failure cases}
\subsection{Examples of human identity evaluations for different face pairs}
\begin{figure}[!hbtp]
    \centering
    {\label{fig:1}\includegraphics[width=0.5\textwidth]{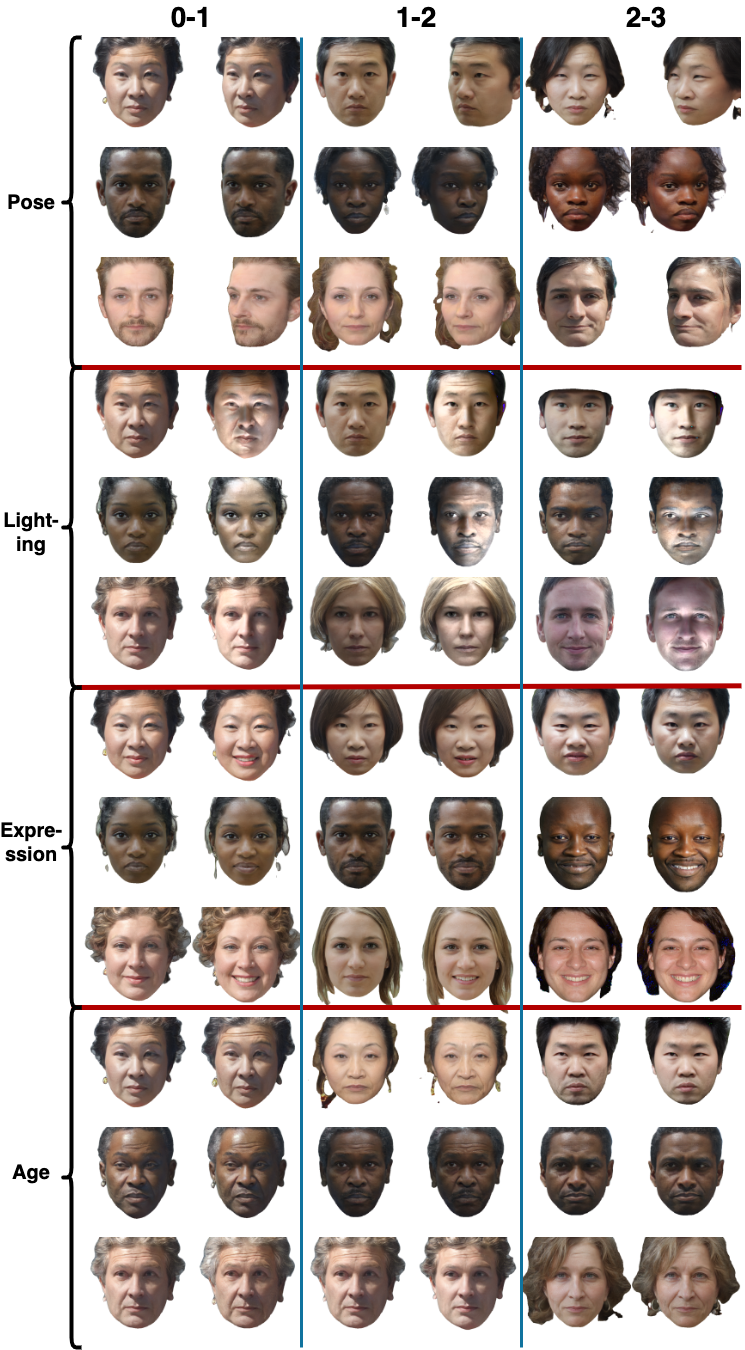}}
    \caption{\textbf{Human identity annotation scores for face pairs intended to belong to the same ID}. All of the shown image pairs are from the same prototype but with non-protected attributes modified to a different degree, as described in Sec. 3. We show the average human annotation scores on top (high score indicates more likely to be from different IDs, raw range is used here $(0-4))$. The last column corresponds to face pairs which humans thought were from different identities, although we intended them to depict the same identity.}
    \label{fig:bins}
\end{figure}
\clearpage
\subsection{Failure case with large ``uncanniness'' score}
\begin{figure}[!hbtp]
    \centering
    {\label{fig:1}\includegraphics[width=0.5\textwidth]{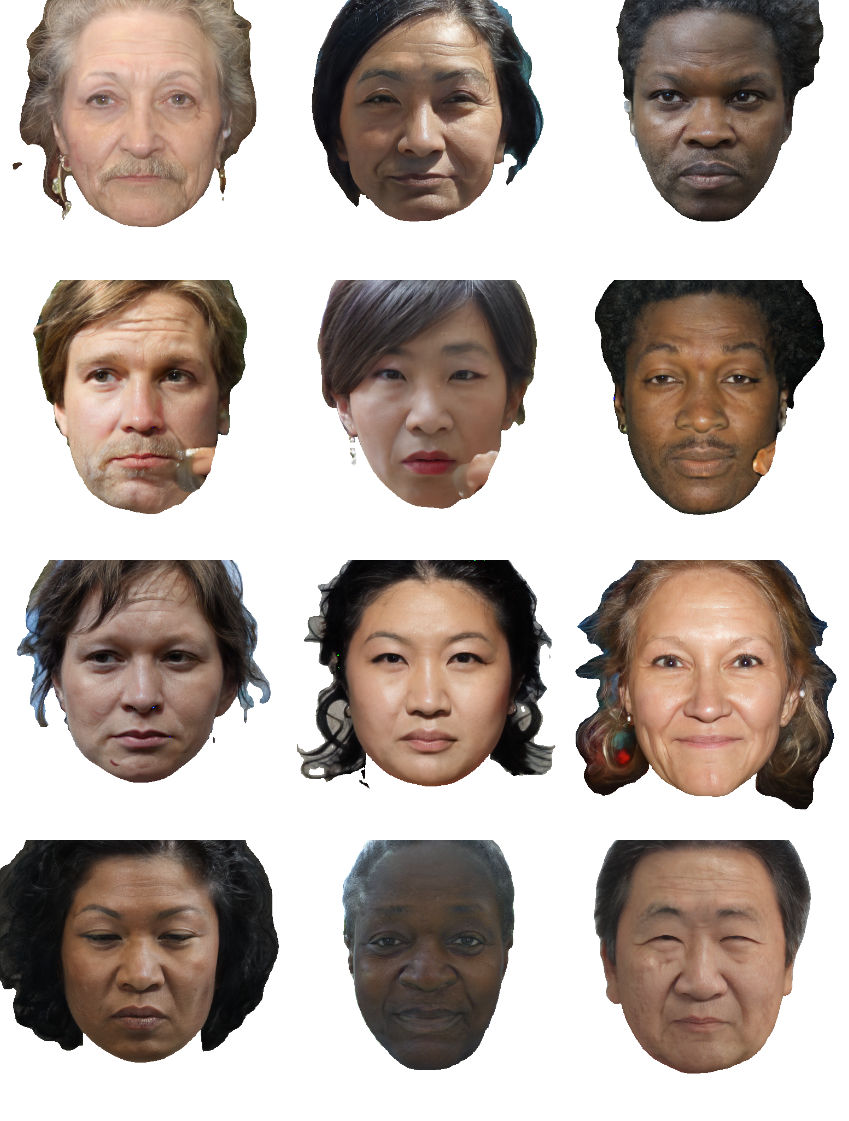}}
    \caption{\textbf{Examples of failure case}. Examples of images whose ``uncanniness'' score are $\geq$ 0.8. There are four main reasons: (a) First row, males with typically female hairstyles. (b) Second row, "ring" artifacts. (c) Third row, bad foreground/background separation makes the hair look unrealistic. (d) Fourth row, other human subjective reasons. We remove all uncanny examples from the test database so that they will not influence experimental results.}
    \label{fig:fail}
\end{figure}

\end{document}